\documentclass[10pt,twocolumn,letterpaper]{article}

\usepackage{wacv} 
\usepackage[accsupp]{axessibility}

\usepackage{graphicx}
\usepackage{amsmath}
\usepackage{amssymb}
\usepackage{booktabs}
\usepackage{multirow}

\usepackage[pagebackref,breaklinks,colorlinks]{hyperref}

\usepackage[capitalize]{cleveref}
\crefname{section}{Sec.}{Secs.}
\Crefname{section}{Section}{Sections}
\Crefname{table}{Table}{Tables}
\crefname{table}{Tab.}{Tabs.}

\def\wacvPaperID{321} 
\def\confName{WACV}
\def\confYear{2025}

\begin{document}

\title{Towards Accurate Unified Anomaly Segmentation}

\author{ \textbf{Wenxin Ma}$^{1,2}$
\qquad 
\textbf{Qingsong Yao}$^{3}$
\qquad 
\textbf{Xiang Zhang}$^{4}$ \qquad 
\textbf{Zhelong Huang}$^{1,2}$
\\
\textbf{Zihang Jiang}$^{1,2}$\footnotemark[1]
\qquad
\textbf{S. Kevin Zhou}$^{1,2,5,6}$\thanks{Corresponding author.} 
\\
 $^1$School of Biomedical Engineering, Division of Life Sciences and Medicine, 
 ~~ \\
 University of Science and Technology of China (USTC), Hefei Anhui, 230026, China
  ~~ \\ 
  $^2$Center for Medical Imaging, Robotics, Analytic Computing \& Learning (MIRACLE), 
  ~~ \\
  Suzhou Institute for Advance Research, USTC, Suzhou Jiangsu, 215123, China
 ~~ \\ 
 $^3$Stanford University, Palo Alto, California, 94025, United State
  ~~ \\ 
 $^4$School of Medicine, Shanghai University, Shanghai 200444, China
  ~~ \\
 $^5$Key Laboratory of Precision and Intelligent Chemistry, USTC, Hefei Anhui, 230026, China
  ~~ \\ 
  $^6$Key Laboratory of Intelligent Information Processing of Chinese Academy of 
  ~~ \\
  Sciences (CAS), Institute of Computing Technology, CAS
 ~~ \\ 
 {\tt\small \{wxma, zhelonghuang\}@mail.ustc.edu.cn jzh0103@ustc.edu.cn }
 ~~ \\
 {\tt\small rookiefcb@shu.edu.cn \{qingsongyao98, s.kevin.zhou\}@gmail.com}
}
\maketitle

\begin{abstract}
\vspace{-3mm}
Unsupervised anomaly detection (UAD) from images strives to model normal data distributions, creating discriminative representations to distinguish and precisely localize anomalies. Despite recent advancements in the efficient and unified one-for-all scheme, challenges persist in accurately segmenting anomalies for further monitoring. Moreover, this problem is obscured by the widely-used AUROC metric under imbalanced UAD settings. This motivates us to emphasize the significance of precise segmentation of anomaly pixels using pAP and DSC as metrics. To address the unsolved segmentation task, we introduce the Unified Anomaly Segmentation (UniAS). UniAS presents a multi-level hybrid pipeline that progressively enhances normal information from coarse to fine, incorporating a novel multi-granularity gated CNN (MGG-CNN) into Transformer layers to explicitly aggregate local details from different granularities. UniAS achieves state-of-the-art anomaly segmentation performance, attaining 65.12/59.33 and 40.06/32.50 in pAP/DSC on the MVTec-AD and VisA datasets, respectively, surpassing previous methods significantly. The codes are shared at \url{https://github.com/Mwxinnn/UniAS}.
\end{abstract}

\vspace{-3mm}
\section{Introduction}
\label{sec:intro}

{\small{
\begin{figure}[t]
 \centering
\includegraphics[width=\columnwidth]{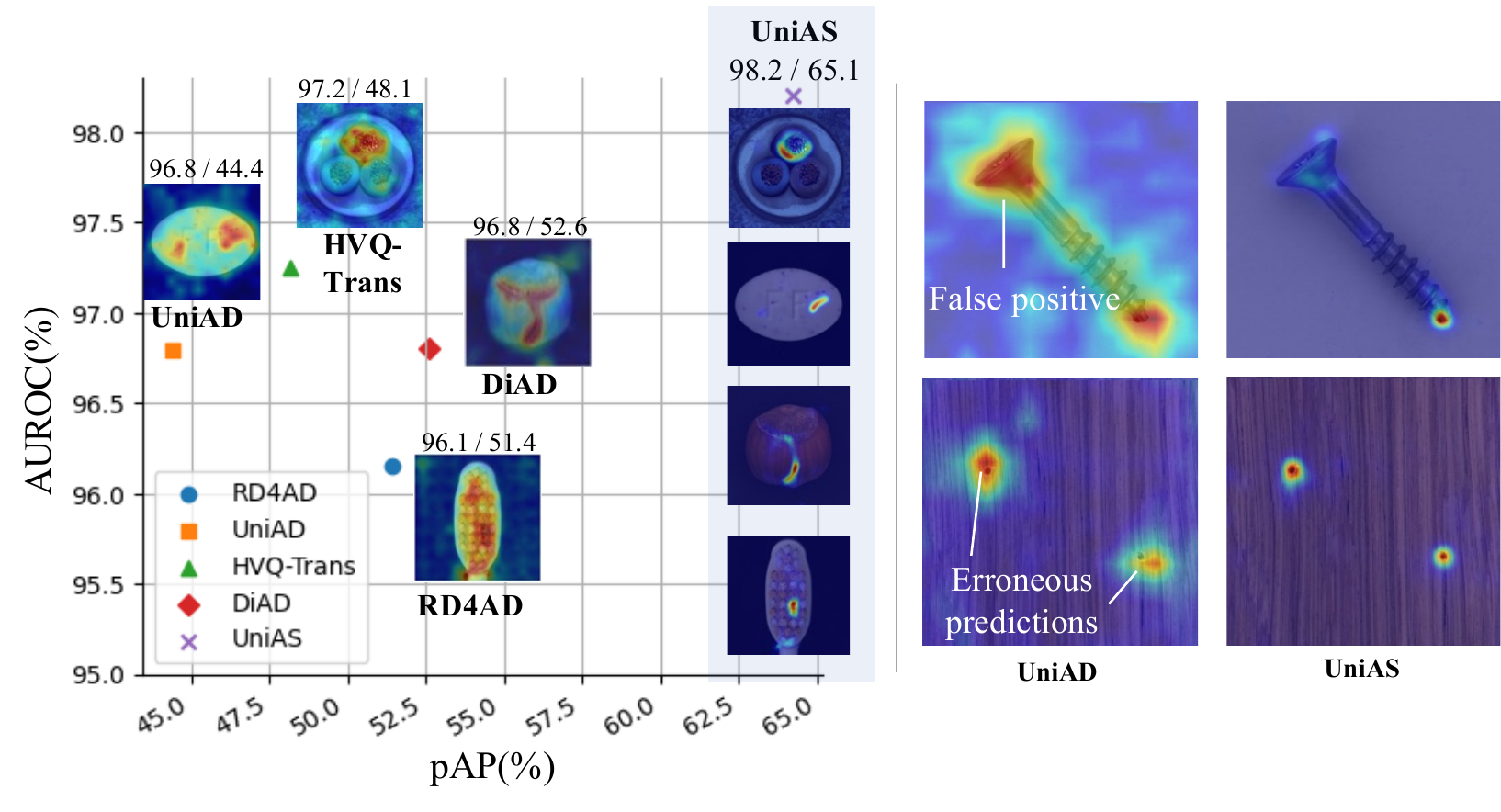}
 \vspace{-9mm}
 \caption{Prediction examples (Left) of the MVTec dataset~\cite{bergmann2019mvtec} and failed cases (Right) of the SOTA model UniAD~\cite{you2022unified}. UniAD has high AUROC but relatively poor segmentation performance, improved by our UniAS.}
 \vspace{-3mm}
 \label{fig:samples}
\end{figure}}}

Unsupervised Anomaly Detection (UAD) in images involves modeling the normal data distribution~\cite{zong2018deep, schlegl2017unsupervised,yao2021label} to detect rare and diverse unexpected signals within visual data, such as lesions in medical images~\cite{fernando2021deep,yao2023adversarial} and defects in industrial images~\cite{bergmann2019mvtec}. While individual solutions per class consume significant computational resources, the newly proposed One-for-All scheme~\cite{you2022unified, lu2023hierarchical, he2023diad, zhao2023omnial} leverages a unified model to capture the complex joint distribution of various normal samples across diverse object classes without the need for fine-tuning. Moreover, extensive studies prove that, compared to traditional One-for-One schemes~\cite{roth2022towards, yao2023focus}, One-for-All UAD methods are faster, more memory-efficient, and show greater promise for generalization~\cite{yao2023one}.

Most recent state-of-the-art (SOTA) UAD frameworks~\cite{you2022unified, lu2023hierarchical, yao2023one, he2023diad} integrate transformer structures to learn to reconstruct compact representations of normal samples that are independent of abnormal samples. This makes it challenging to reconstruct defect-corrupted representations. Accordingly, the difference between reconstructed normal and abnormal features indicates anomalies. These frameworks achieve significant progress as assessed by the widely used metric, the area under the receiver operating characteristic curve (AUROC).

Despite significant advancements in image-level anomaly detection, recent observations~\cite{bertoldo2024aupimo, rafiei2023pixel} indicate that even the top-performing models still struggle to precisely segment anomalies, often resulting in false-positive (FP) predictions (as shown in \cref{fig:samples}). Due to the class imbalance caused by numerous small anomalies in AD settings, AUROC tends to be a less effective indicator of model performance. High AUROC scores reported in the literature can possibly overlook predicted false-positive (FP) pixels, especially when true negative (TN) background pixels constitute the majority~\cite{rafiei2023pixel}. Nonetheless, precisely segmenting abnormal pixels is significant for quantifying the degree of anomaly and facilitating subsequent anomaly monitoring and modification~\cite{zhang2023destseg}.

In this paper, we delve deeper into the unresolved One-for-All anomaly segmentation challenge. We first demonstrate the inadequacy of solely relying on AUROC as the metric for evaluating anomaly segmentation and involve metrics such as pixel-wise Average Precision (pAP) and Dice Similarity Coefficient (DSC), which are widely used to evaluate segmentation performance. Accordingly, we find that the original cutting-edge feature-reconstruction-based One-for-All UAD methods show unsatisfactory performance on these segmentation metrics. We argue that the problem stems from their downsampling operations on the CNN-extracted features, which are then reconstructed at a coarse scale~\cite{you2022unified, lu2023hierarchical, yao2023one}, compromising the intricate low-level information crucial for precise anomaly segmentation.

Motivated by this observation, we introduce our Unified Anomaly Segmentation (UniAS) approach, which considers multi-level features to enhance segmentation performance. UniAS establishes a multi-level hybrid pipeline to gradually reconstruct the extracted features in a level-by-level manner. Within each level, we propose a hybrid transformer-CNN module that leverages the transformer's global receptive field to capture high-level features of normal data while simultaneously utilizing our Multi-Granularity Gated CNN (MGG-CNN) to recover intricate local details of normal samples. When anomalies are present, transformers can semantically detect anomalous pixels, with MGG-CNNs helping delineate the granularity-relevant boundaries of anomalies. By aggregating anomaly signals at different levels and granularities from coarse to fine, UniAS leverages both high and low-level anomaly maps to suppress semantic false-positive regions and manage to precisely segment anomalies eventually.

We conduct extensive experiments to demonstrate the effectiveness of our multi-level hybrid pipeline. Comparing UniAS with various cutting-edge models, our model shows significant improvement in precisely segmenting anomalies, as shown in \cref{fig:samples} (Left). Our model achieves the highest pAP with excellent AUROC, and FP predictions are significantly suppressed by our design, as shown in \cref{fig:samples} (Right).

In summary, we carefully analyze the ineffectiveness of AUROC under imbalanced situations and propose our UniAS to improve the performance of one-for-all anomaly segmentation. Our primary contributions are:

\begin{enumerate}
\item We explore the \textbf{limitation of AUROC} as the sole metric and discover the underlying FP problem in anomaly segmentation. Accordingly, we suggest greater emphasis on additional metrics including \textbf{pAP and DSC} for the evaluation of anomaly segmentation task.
\item We propose our new model, UniAS, which follows a \textbf{multi-level hybrid pipeline}, leveraging the benefits of the transformer's global properties and the CNN's locality to foster a holistic understanding of features across multiple levels and granularities, facilitating precise anomaly segmentation.
\item Extensive experimental results on widely used MVTec-AD and VisA datasets demonstrate the effectiveness of UniAS, setting new SOTA to 65.12/59.33 and 40.06/32.50 in pAP and DSC, respectively, surpassing previous models by a large margin.
\end{enumerate}

\section{Related Work}
\label{sec:RelatedWork}
\subsection{Unsupervised Anomaly Detection}
Current methods can primarily be divided into the following three categories: 1) \textbf{Augmentation-based methods}~\cite{li2021cutpaste, zavrtanik2021draem, liu2023simplenet, zhang2021defect, zhang2023prototypical} typically involve the insertion of artificially generated anomalous patterns into normal signals, transforming the task into a supervised task. High-quality synthetic anomalies must be ensured. 2) \textbf{Boundary-based discriminative methods} compactly map normal features into a higher-dimensional space that separates them from abnormal components. To enhance discriminative capabilities, specialized designs~\cite{liu2023simplenet, defard2021padim, roth2022towards,gudovskiy2022cflow, rudolph2022fully, kim2023sanflow, jimenez2015variational,deng2022anomaly, zhang2023destseg, wang2021student,wang2023uncertainty} need to be developed, which can be computationally expensive. 
3) \textbf{Reconstruction-based methods} train a model on normal samples exclusively~\cite{chen2022utrad, bergmann2018improving, yao2023one, he2023diad, yao2023focus}, which is reconstructed in either feature or image space, with the assumption that the reconstruction error is notably higher for anomalous inputs. This approach is suitable for One-for-All settings~\cite{you2022unified, he2023diad, lu2023hierarchical, yao2023one}, where the distribution of either normal data or anomalies is highly complex, making it difficult to delineate a classification boundary or synthesize artificial anomaly samples to aid training. 
However, the reconstruction model may have ``identical shortcut'' issue where abnormal inputs are also well reconstructed~\cite{you2022unified, you2022adtr, lu2023hierarchical}. Additionally, normal images may contain details that are challenging to reconstruct, resulting in significant false positive predictions. The majority of current methods utilize downsampling techniques to highlight differences in high-dimensional semantics, neglecting lower-level features that are crucial for anomaly segmentation. 

\subsection{Transformers in Pixel-Level Tasks}
The Transformer architecture has proven effective in various pixel-level tasks, including object detection and segmentation~\cite{carion2020end, cheng2021per, cheng2022masked, jain2023oneformer}. By reformulating the tasks as a target-querying problem, they utilize learnable queries to encapsulate semantic information within features extracted by CNNs, ultimately used to match targets as bounding boxes~\cite{carion2020end} or segmentation masks~\cite{cheng2021per, cheng2022masked}. Specifically, DETR and MaskFormer process features at the highest level, while Mask2Former~\cite{cheng2022masked} sequentially processes features at each scale, achieving widespread success in segmentation~\cite{kirillov2019panoptic}.

For reconstruction-based One-for-All UAD, recent works~\cite{you2022unified, lu2023hierarchical} have endowed the queries with a new role as a memory matrix to help reconstruct normal samples. UniAD~\cite{you2022unified} compresses the features to the same resolution and uses a MaskFormer-like structure. However, this approach fails to capitalize on the advantages of multi-level processing as Mask2Former. Our UniAS maximizes the benefits of a Mask2Former-like structure, with one-for-all AD-targeted modifications and achieving SOTA segmentation performance in UAD.
\section{Our Approach}
\subsection{Overview}
\subsubsection{Problem Formulation} 
One-for-All anomaly segmentation is defined as follows: Consider a set of anomaly-free data $X^{train}=\{x_1,x_2,...,x_n\}$ across various categories, and a test set $X^{test}=\{x'_1,x'_2,...,x'_n\}$ comprising both normal and abnormal samples from all the above categories. The general objective is to train a unified model capable of accurately identifying various anomalous pixels in anomaly samples across different categories within the test set. For reconstruction-based UAD, the training process takes anomaly-free data $X^{train}$ as input, learning to replicate the distribution of the data or its features, while during inference, the model outputs pixel-wise reconstruction error as anomaly map $A$. This map is taken as the segmentation prediction accordingly.

\subsubsection{Observation: The limitation of AUROC}
\label{sec:problems}
 The Area Under the Receiver Operating Characteristic Curve (AUROC) between the anomaly map $A$ and the segmentation mask has been the most widely used metric for AD in pixel level~\cite{li2021cutpaste, lin2017feature,defard2021padim, chen2022utrad, deng2022anomaly, gudovskiy2022cflow, he2023diad, you2022unified, lu2023hierarchical, liu2023simplenet, roth2022towards, xiang2023squid, yao2023one, you2022adtr, zhang2023destseg, zhang2023prototypical}, which is calculated as \cref{eq:AUROC}. 
 Here, TP stands for correctly predicted anomaly pixels, while TN stands for correctly predicted normal pixels. FP and FN are false-positive and false-negative pixels, respectively.
 \begin{equation}
 \begin{aligned}
      \text{AUROC} = \int_{0}^{1} &\text{TPR}\ d(\text{FPR}),\\
      ~~\text{TPR}= \frac{\text{TP}}{\text{TP}+\text{FN}},
      ~~&\text{FPR}=\frac{\text{FP}}{\text{FP}+\text{TN}}.
 \end{aligned}
\label{eq:AUROC}
 \end{equation}

{\small{
\begin{figure}[t]
    
    \includegraphics[width=\linewidth]{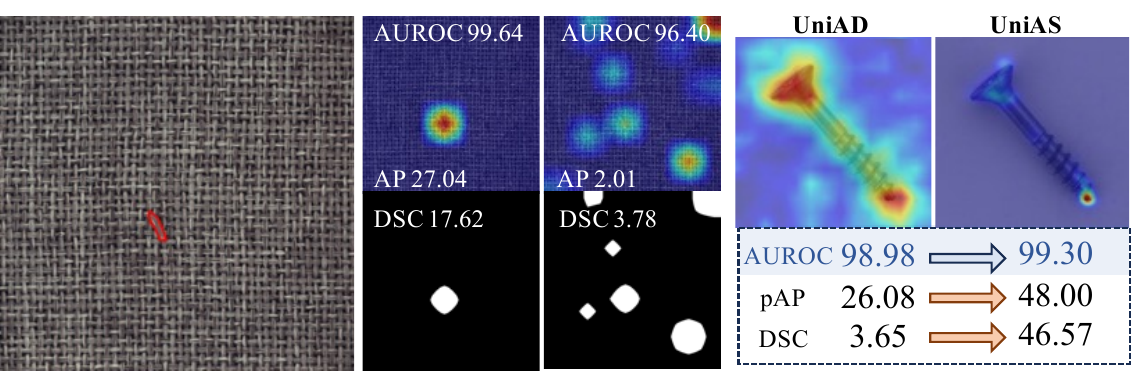}
    \vspace{-6mm}
    \caption{(Left) An example of MVTec-AD with a red line delineating anomalous GT~\cite{bergmann2019mvtec}, along with toy anomaly maps, segmentation predictions, and the corresponding metrics. Bad predictions can have high AUROC numbers. (Right) A real prediction example and corresponding metrics, showing the limitation of AUROC.}
    \label{fig:toy}
    \vspace{-3mm}
\end{figure}}}
The performances reported in the literature are converging towards 100\% regarding pixel-level AUROC, 
giving the impression that the pixel-level AD task has been solved~\cite{bertoldo2024aupimo}. 
However, anomaly samples typically appear rarely with few anomalous pixels~\cite{bertoldo2024aupimo, rafiei2023pixel} (as shown in Fig.~\ref{fig:toy} (Left)). 
The imbalance problem can negatively impact the AUROC's effectiveness, leading to failures in extreme scenarios~\cite{saito2015precision}. 
As shown in \cref{eq:AUROC}, when detecting small anomaly signals, TN tends to be very large, resulting in a low FPR regardless of FP predictions. 
At the same time, the upsampling operation in recent works~\cite{you2022unified,lu2023hierarchical} ensures a high TPR with erroneous predictions (high FP). 
These two factors contribute to the inflated AUROC numbers. 
We generate toy predictions as an example to demonstrate this serious problem of AUROC in \cref{fig:toy} (Left). AUROC tends to be very high even with unsatisfactory segmentation results. 

For real situations, we calculate the percentage of anomalous pixels in MVTec and VisA, denoted as Anomaly Rate (AR), to reference the degree of imbalance in the datasets. A lower AR indicates that anomalous samples mainly consist of small regions, which can possibly lead to the discussed problem. The average AR in both datasets is relatively low (13.17\% and 3.00\%, respectively). As shown in \cref{fig:toy} (Right). UniAD's prediction has a high AUROC value with noticeable FP, while other metrics focusing on TP, such as pAP and DSC, can better reflect the precision of the result by eliminating the TP-TN imbalance problem. 

In summarize,
our observation indicates that high performance in the AUROC metric alone does not necessarily correlate with effective anomaly segmentation.

\subsubsection{Motivation and Overall Design}

\textbf{Motivation:} As demonstrated, pixel-level segmentation remains an unsolved task. We attribute this to downsampling in recent methods and the consequent loss of low-level features, which are crucial for capturing intricate details in anomaly segmentation. This motivates us to propose UniAS, aiming to model and aggregate precise multi-scale features to enhance pixel-level segmentation performance.

\textbf{The Overall Architecture} of our UniAS is depicted in \cref{fig:main}. 
Initially, multi-scale features are extracted by a CNN backbone (see \cref{sec:extraction}) which serves as the inputs and targets of reconstruction. 
To preserve information from different levels, we propose a \textbf{multi-level pipeline} to reconstruct features hierarchically. 
To precisely model the distribution of normal samples, we use a \textbf{hybrid architecture} combining Transformer and Multi-Granularity Gated CNN (MGG-CNN) together, leveraging their global and local benefits.
 In particular, MGG-CNN is designed to amalgamate details from diverse granularities. The whole process is aided by a Sample-Aware Reweighted (SAR) Query to enhance One-for-All ability (see \cref{sec:embedding}).
Finally, the anomaly maps (\cref{sec:loss}) are produced by comparing reconstructed features with extracted ones from various levels, then pixel-wisely multiplied together to form the final prediction. 
\label{sec:Method}
{\small{
\begin{figure*}[t]
    \centering
    \includegraphics[width=0.95\textwidth]{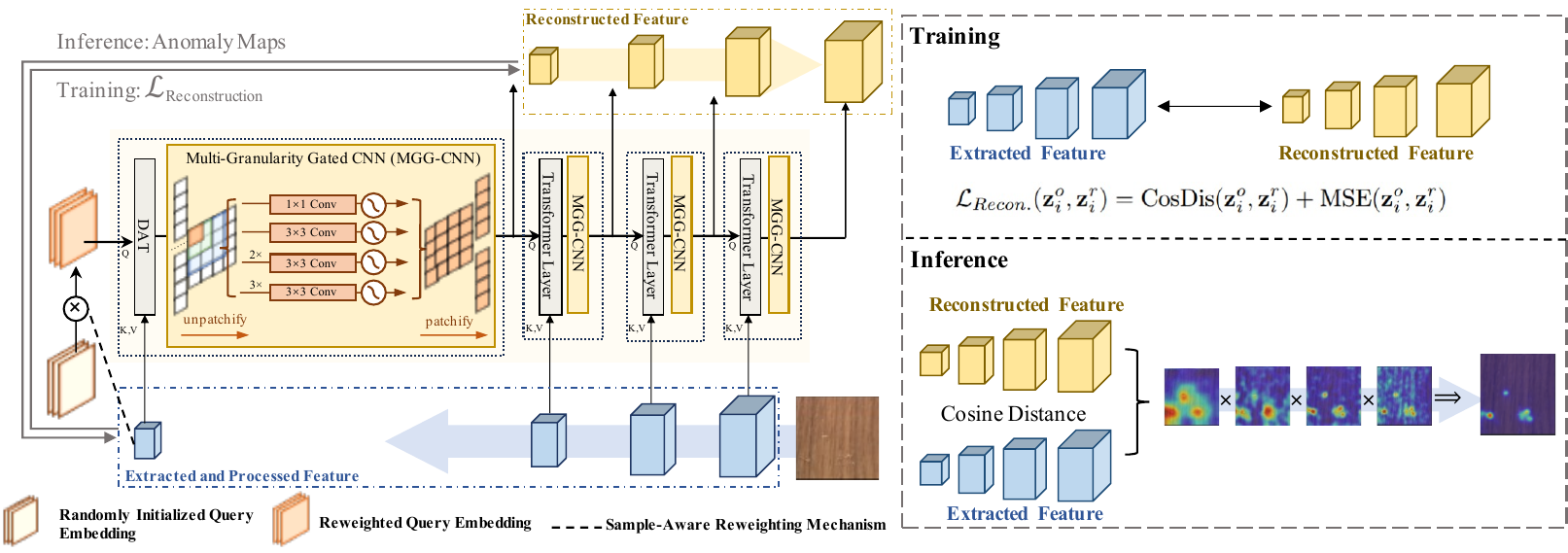}
    \vspace{-3mm}
    \caption{\textbf{(Left) The overview of UniAS.} After extracting features, our multi-level hybrid decoder, composed of Transformer and MGG-CNN hybrid blocks, hierarchically reconstructs normal features from coarse to fine. Transformers and MGG-CNNs play complementary roles in global and local modeling. The SAR Query is incorporated to facilitate One-for-All segmentation.  \textbf{(Right) Illustration of the training loss and anomaly map generation during inference.} By multiplying anomaly maps together, information from separate levels is aggregated, facilitating accurate anomaly segmentation.}
    \vspace{-3mm}
    \label{fig:main}
\end{figure*}}}

\subsection{Feature Extraction and Filtering}
\label{sec:extraction}
 For a normal input image $x_i\in \mathbb{R}^{H\times W\times 3}$, we use an ImageNet-pretrained CNN with $N$ pyramidal structures $\{\Phi_1,\Phi_2,...\Phi_N\}$ to extract $K$ multi-scale features, denoted by $\mathcal{Z}^{ori}=\{\mathbf{z}_1^o, ...,\mathbf{z}_K^o | \mathbf{z}_i\in \mathbb{R}^{H_i\times W_i\times C_i}\}$, where $H_i$, $W_i$, and $C_i$ represent the height, width, and channel size of the $i$-th feature map. Additionally, as demonstrated to be effective in~\cite{roth2022towards, liu2023simplenet}, we apply a Gaussian filter $W_{G}$ to aggregate the neighbor information and concatenate the residual to preserve details:
\begin{equation}
\begin{aligned}
      &\mathcal{Z}^{ori}=\{\mathbf{z}_1^o, ...,\mathbf{z}_K^o|\mathbf{z}_i^o=f_{\Phi_i}(x) \odot W_{G}\}, \\
      & \mathcal{Z}^{cat}=\{\mathbf{z}_1^c, ...,\mathbf{z}_K^c|\mathbf{z}_{i}^c=cat(\mathbf{z}_i^o,\mathbf{z}_i^o-f_{\Phi_i}(x))\}. 
\end{aligned}  
\label{eq:feature}
\end{equation}

\subsection{Multi-level Hybrid Reconstruction Pipeline}
\label{sec:Recon}
 After extracting $K$ levels of feature, UniAS learns to model and reconstruct the features at each level. Each reconstruction block consists of a transformer layer for \textbf{global} modeling and a MGG-CNN for \textbf{local} refinement. The transformer layer introduces normal signals to MGG-CNN, aided by a SAR query which acts as a memory matrix at the beginning (see details in \cref{sec:embedding}). 

\subsubsection{Normal Data Modeling} 

\textbf{Patchify process:} Given the varying resolutions of the extracted features and computational constraints, we ensure the number of patches at each level is consistent. This is achieved by utilizing various patch embedding layers ${\theta_1,...,\theta_k}$ to divide the features $\mathcal{Z}^{input}$ into patches with different sizes. Specifically, for the $i$-th feature $\mathbf{z}_i \in \mathcal{Z}^{input}$ with resolution $H_i \times W_i$, the patch size is determined as $H_i/H_K \times W_i/W_K$, where $H_K$ and $W_K$ are the height and width of the feature at the highest $K$-th level, resulting in $N=H_K \times W_K$ patches. These $N$ embeddings are denoted as: 
\begin{equation}
    \mathbf{h}_i=f_{\theta_i}(\mathbf{z}_i),~~\mathbf{h}_i \in \mathbb{R}^{N\times C'},~~i=1,2,...,K
\end{equation}

\noindent{\textbf{Global Modeling:}}
Transformer’s global receptive field allows it to capture long-range dependencies and context across the entire input, enabling more comprehensive and accurate modeling of complex patterns and relationships~\cite{vaswani2017attention,carion2020end,cheng2021per,cheng2022masked}. To take advantage of this, we use transformer layers similar to Mask2Former~\cite{cheng2022masked}, represented as:
\begin{equation}
    \begin{aligned}
        &\mathbf{d}_{CA} = \text{MHCA}(\mathbf{Q}, \mathbf{K}, \mathbf{V})+\mathbf{Q},\\
        &\mathbf{d}_{SA} = \text{MHSA}(\mathbf{d}_{CA})+\mathbf{d}_{CA},\\
        &\mathbf{d} = \text{FFN}(\mathbf{d}_{SA})+\mathbf{d}_{SA},\\
    \end{aligned}
    \label{eq:attn}
\end{equation}
where $\text{MHCA}(\cdot)$, $\text{MHSA}(\cdot)$, and $\text{FFN}(\cdot)$ are the standard multi-head cross attention, multi-head self-attention and feed-forward network in a vanilla Transformer~\cite{dosovitskiy2020image}. It takes the SAR query $\mathbf{q}_0$ as the initial $\mathbf{Q}$, which is then updated by the output of $(i-1)$-th layer $\mathbf{q}_i$, and it takes feature patches $\mathbf{h}_{K-i}$ as $\mathbf{K}$ and $\mathbf{V}$ for calculating attention, from high level to low level.
\vspace{1.5mm}
\noindent{\textbf{Local Modeling:}}
The CNN's local receptive field and inherent inductive bias excel at capturing fine-grained details and local patterns, making it highly effective for local feature extraction. This capability complements the transformer's global modeling strength, creating a synergistic effect where the CNN handles local refinements while the transformer captures broader context \cite{cao2022swin,chen2021transunet,liu2021swin,tang2023mobileutr}. 

Considering the diverse granularities of anomalies across the dataset, we propose Multi-Granularity Gated CNN (MGG-CNN), by employing multiple branches with different receptive fields for better multi-granularity modeling. As depicted in \cref{fig:main} (left), these branches consist of one $Conv1\times1$ branch accompanied by 3 $Conv3\times3$ branches, where the $3\times3$ convolutional block is stacked for 1,2 and 3 times, respectively. This design creates an MGG-CNN with a dynamic combination of receptive fields spanning 1, 3, 5, and 7. Furthermore, a GELU layer~\cite{hendrycks2016gaussian} acts as a gate function to activate the feature of each branch to encourage distinct information at each scale and introduce more non-linearity. Each branch's output is added together. Our MGG-CNN is expressed as:
\begin{equation}
\mathbf{q}_i=\text{Gate}(Conv1(\mathbf{d}_i))+\sum_{j=1}^{3}\text{Gate}(Conv3^{\ j-1}(\mathbf{d}_i)),
\end{equation}
where each $\mathbf{d}_i$ is the output of last transformer layer, and $\mathbf{q}_i$ is the input for the next level transformer layer for multi-level modeling.

\subsubsection{Reconstruction and Anomaly Maps}
\label{sec:loss}
Following each MGG-CNN, an unpatchify operation and a transposed convolution layer are applied, resulting in the final reconstructed features $\mathcal{Z}^{rec}=\{\mathbf{z}^r_1, \mathbf{z}^r_2,...,\mathbf{z}^r_K\}$ with the same dimensions as extracted feature $\mathbf{z}_i^{ori}$.  
 As shown in \cref{fig:main}, during training, the loss is calculated as the sum of Cosine Distance (CosDis) and Mean Squared Error (MSE), following \cite{salehi2021multiresolution}.
 During inference, we employ cosine distance to measure the dissimilarity between the reconstructed features and extracted features at each level, resulting in $K$ anomaly maps. These maps are then upsampled to the same size $H\times W$ as input images.  The final prediction $A$ is generated by multiplying these maps together, as \cref{eq:anomaly}. Based on these designs, UniAS can accurately segment anomalies in a coarse-to-fine fashion. 
\begin{equation}
A=\prod_{i=1}^{K}\text{upsample}(1-\frac{{<\mathbf{z}^r_i, \mathbf{z}^o_i>}}{{\|\mathbf{z}^r_i\| \cdot \|\mathbf{z}^o_i\|}}).
\label{eq:anomaly}
\end{equation}

\subsection{Sample-Aware Reweighting Mechanism For One-for-all Segmentation}
\label{sec:embedding}
In addition to improving segmentation performance, we have also incorporated a Sample-Aware Reweighting (SAR) mechanism to initialize more sample-specific accurate prototypes as queries to boost One-for-All performance. 

\vspace{1.5mm}
\noindent{\textbf{Motivation:}} Features extracted from various samples, particularly those from distinct categories, often demonstrate substantial disparities in high-level characteristics. Employing a shared learnable query to memorize patterns across all categories can be suboptimal for this variability. We argue that incorporating sample-specific information into the query can help One-for-All reconstruction.

\vspace{1.5mm}
\noindent{\textbf{Design:}} Before training, the query $\mathbf{q}_0^{ori}\in \mathbb{R}^{ H_K\times W_K\times C'}$ is randomly initialized, whose shape is the same as the highest patch embedding $\mathbf{h}_K$. We utilize channel and spatial weights in CBAM~\cite{woo2018cbam} to reweight query $\mathbf{q}_0^{ori}$ based on $\mathbf{h}_K$ with the richest semantic. The multiplied $\mathbf{q}_0\in  \mathbb{R}^{ H_k\times W_k\times C'}$ is the query of the first transformer layer. This reweighted query incorporates sample-specific information, making it distinct and better suited to the variability across samples (see in Supplementary Material for details).
\section{Experiments}
\begin{table*}[t]
\centering

\resizebox{\textwidth}{!}{%
{\small{
\begin{tabular}{l|c|c c c|c c c c c|c}
\toprule
\multirow{2}{*}{Dataset} & \multirow{2}{*}{\color{blue}AR(\%)}  &\multicolumn{3}{c|}{(Originally) One-for-One}&\multicolumn{6}{c}{One-for-All} \\
\cmidrule(lr){3-5}\cmidrule(lr){6-11}
&  &SimpleNet\cite{liu2023simplenet}&RD4AD\cite{deng2022anomaly} & DeSTSeg\cite{zhang2023destseg} & UniAD\cite{you2022unified} &OmniAL~\cite{zhao2023omnial}& HVQ-Trans\cite{lu2023hierarchical} & DiAD~\cite{he2023diad}& ViTAD~\cite{zhang2023exploring} & \textbf{UniAS(Ours)} \\
\midrule
    MVTec&{\color{blue}13.17}& 96.78 &96.15&94.41&96.79({96.80*}) &\textbf{{98.30*}}&97.25({97.30*})&{96.80*}&{97.70*}&98.22 \\
VisA&{\color{blue}3.00}& 96.82 &97.04&94.28&98.36&{96.60*}&98.41({\textbf{98.70*}})&{96.00*}&{98.20*}&97.80 \\
\bottomrule
\end{tabular}
}}}
\vspace{-3mm}
\caption{\textbf{The performance of recent SOTA models in pixel-AUROC(\%)}. Results with * are from the original paper, and other results are obtained from our reproduction. Best results are \textbf{bold}. Works are chronically arranged.}
\label{tab:auc}
\vspace{-2mm}
\end{table*}

\subsection{Metrics}
 As discussed in \cref{sec:problems}, we analyze Anomaly Rate (AR) to show the degree of imbalance in the datasets. Accordingly, we adopt pAP and DSC apart from AUROC for anomaly segmentation evaluation. pAP evaluates the average precision across different recall levels, prioritizing successfully detecting positive pixels and keeping robust under class imbalance~\cite{zavrtanik2021draem, zhang2023destseg, he2023diad}. Meanwhile, DSC, a common metric in segmentation tasks, demonstrates robustness regardless of object size~\cite{hu2023label, yao2021label}. In our implementation, we \textit{determine the segmentation threshold by maximizing the sum of precision and recall on the Precision-Recall (PR) curve for each category.}

\subsection{Datasets}
\subsubsection{MVTec-AD~\cite{bergmann2019mvtec}}
MVTec-AD is a widely-used industrial anomaly detection dataset of real-world scenarios. It comprises 15 classes, totaling 5,354 high-resolution images spanning various categories, with one object in an image and a clean background. Each class consists of normal training samples, while the test set includes both normal and anomalous samples. For each anomalous sample in the test set, pixel-level ground-truth annotations are provided for segmentation evaluation. 

\subsubsection{VisA~\cite{zou2022spot}} 
VisA is a recent industrial anomaly detection dataset containing more challenging scenarios. This dataset comprises a total of 10,821 images, covering 12 object categories with 78 different types of anomalies, including scratches, dents, color spots, cracks, and structural defects. Images in the dataset exhibit complex structures, objects positioned in various locations, with tiny anomalous sizes and noisy backgrounds.

\subsection{Implementation Details}
 All images in MVTec-AD and VisA are resized to $224\times 224$. We adopt a pretrained and fixed EfficientNet~\cite{tan2019efficientnet} to extract features from stage-1 to stage-4. Channel dimension $C'$ for patch embedding layers is 256. AdamW~\cite{loshchilov2017decoupled} with weight decay 0.0001 is adopted for optimization. Our UniAS decoder is trained from scratch for 1000 epochs on a RTX 3090 GPU  with batch size 64. The learning rate is initially set to $1\times 10^{-4}$, and reduced by a factor of 0.1 every 400 epochs. For more details, please refer to the Supplementary Material.

\subsection{Limitation of AUROC}
We first present our results in AUROC in \cref{tab:auc}, compared with recent works. It can be observed that recent works show a noticeable saturation trend in reported AUROC across both datasets. However, it is important to note that the AUROC scores reported in the literature may be inflated and misleading for evaluating segmentation performance due to the imbalanced distribution between normal and abnormal pixels, as indicated in the {\color{blue}AR} column. Specifically, in the VisA dataset, which has a notably low anomaly rate, HVQ-Trans shows the highest AUROC values. Despite this, a substantial segmentation performance gap exists, as shown in the segmentation visualization in \cref{fig:results}. Consequently, we shift our focus to other metrics such as pAP and DSC in the following discussion. Detailed AUROC results are provided in the Supplementary Material.

\subsection{Qualitative Results}
{\small{
\begin{figure*}[!ht]
    \centering
    \includegraphics[width=0.9\textwidth]{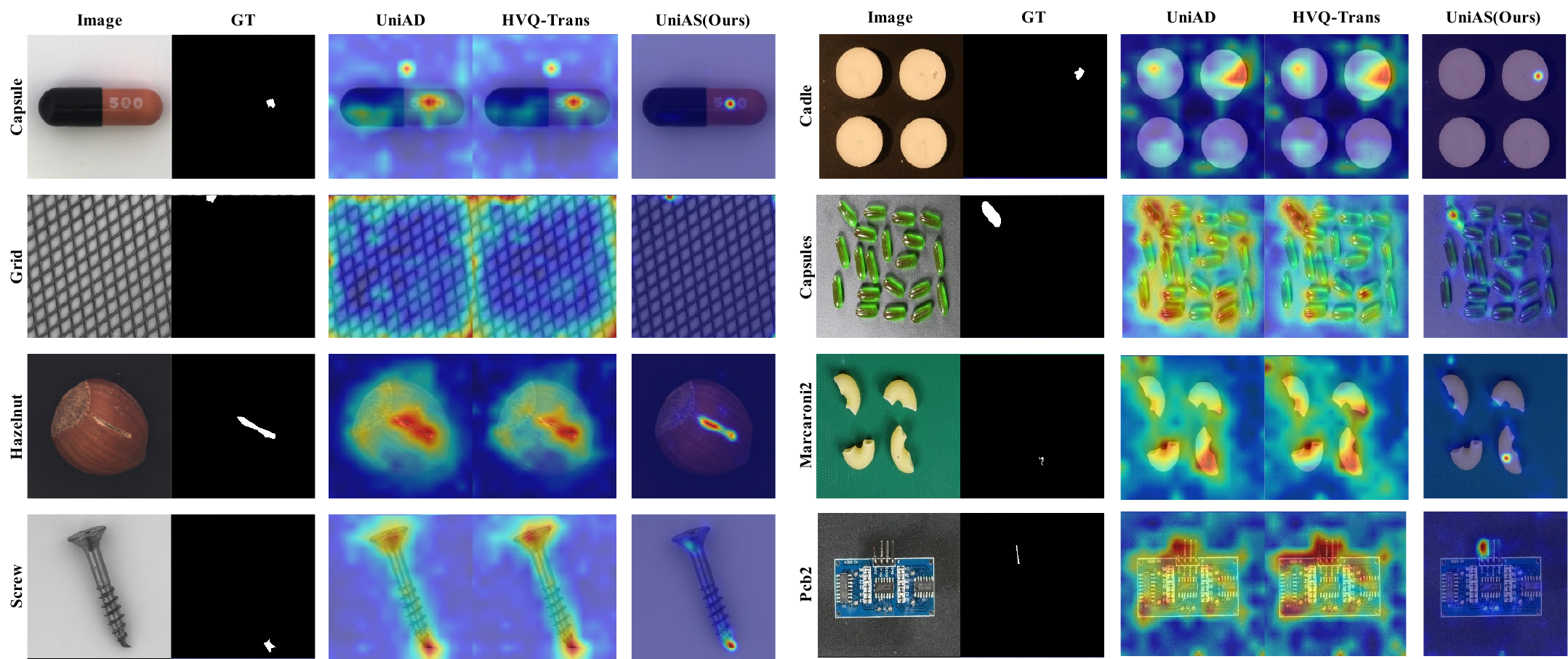}
    \vspace{-1.5mm}
    \caption{\textbf{Qualitative results on MVTec-AD(left) and VisA(right).} We visualize the anomaly maps and the generated masks. UniAS localizes anomaly precisely, exhibiting meaningful segmentation results.}
    \label{fig:results}
    \vspace{-3mm}
\end{figure*}}}
{\small{
\begin{figure*}[!h]
    \centering
    \includegraphics[width=0.8\textwidth]{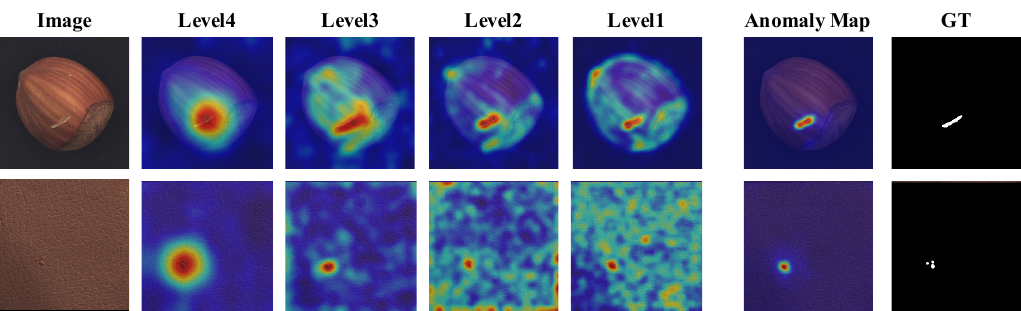}
    \vspace{-2mm}
    \caption{(Left) Examples of input images and corresponding anomaly maps from feature level 4 to level 1. (Right) Aggregated anomaly maps and the anomaly ground truth masks. Anomaly maps play complementary roles, detecting anomalous pixels from coarse to fine.}
    \label{fig:multi-level}
\end{figure*}}}

 We visualize the results from both datasets in \cref{fig:results}, comparing with two recent One-for-All SOTA models, UniAD\cite{you2022unified} and HVQ-Trans~\cite{lu2023hierarchical}. While they can successfully distinguish anomalies, they exhibit numerous false positive predictions (\eg, Screw, Candle, and Capsule). In contrast, our design demonstrates more accurate segmentation performance across various anomaly categories. Particularly, UniAS mitigates erroneously FP predictions, thereby achieving better overall segmentation precision and offering valuable assistance in anomaly monitoring and removal.

We attribute the improvement to our design of multi-level feature reconstruction, which successfully model features at multiple levels, leveraging their complementary roles to suppress false predictions. In particular, higher-level features concentrate on semantics, facilitating accurate overall anomaly localization, while lower-level features have rich textural information aiding in delineating the shape of the anomalous region. Multiplying them together maximizes the complementary benefits, ensuring a precise segmentation result. We also visualize the anomaly maps of each layer, as depicted in \cref{fig:multi-level}. We label four levels of EfficientNet-extracted features 1 to 4, from the lowest to the highest. As shown, while each layer's anomaly map consistently highlights the anomalous areas, their respective roles differ, contributing to the eventual precise result. 

\subsection{Quantitative Results}
We choose seven of the best SOTA models to compare. These models include a boundary-based approach with a feature memory bank incorporated, \textbf{PatchCore}~\cite{roth2022towards}; an image reconstruction-based method with pseudo-anomalies, \textbf{DRAEM}~\cite{zavrtanik2021draem}; an image reconstruction-based method with diffusion, \textbf{DiAD}~\cite{he2023diad}; and four feature reconstruction-based methods, namely, \textbf{DeSTSeg}~\cite{zhang2023destseg} and \textbf{RD4AD}~\cite{deng2022anomaly} with CNN structures, as well as \textbf{UniAD}~\cite{you2022unified} and \textbf{HVQ-Trans}~\cite{lu2023hierarchical} with Transformer structures. UniAD and HVQ-Trans are two closely related transformer-based One-for-All works, while the remaining methods are originally designed as one-for-one approaches. For a fair comparison, we adapt these methods within the One-for-All scheme with the special designs unchanged. We re-trained the models using their publicly available official code with the default implementation for measuring pAP and DSC.

\begin{table}
  \centering
  \resizebox{0.9\columnwidth}{!}{%
  {\small{
  \begin{tabular}{@{}c c|c c c @{}}
    \toprule
    \multicolumn{2}{c}{Method} & MVTec & VisA & Mean \\
    \midrule
    PatchCore & {\color{blue} 22'CVPR} & 55.73 & 35.64 & 45.69 \\
    DRAEM & {\color{blue}21'ICCV} & 49.86 & 34.49 & 42.18 \\
    DeSTSeg &{\color{blue}23'CVPR} & 59.20 & 33.35 & 46.28 \\
    RD4AD & {\color{blue}22'CVPR} &51.41 & 38.23 & 44.82 \\
    UniAD &{\color{blue}22'NIPS} & 44.42 & 34.32 & 39.37 \\
    HVQ-Trans&{\color{blue}23'NIPS} & 48.15 & 33.17 & 40.66 \\
    DiAD&{\color{blue}24'AAAI} & {52.60*} & {26.10*} & 39.35* \\
    \midrule
    UniAS (Ours)&{\color{blue}25'WACV} & \textbf{65.12} & \textbf{40.06} & \textbf{52.29}\\
    \bottomrule
  \end{tabular}
  }}}
  \vspace{-2mm}
  \caption{{Quantitative Results on MVTec-AD and VisA in pAP(\%)}. The best results are \textbf{bold}, and results with * are from the original paper.}
  \label{tab:pAP}
  \vspace{-1mm}
\end{table}

\begin{table}
  \centering
  \resizebox{0.9\columnwidth}{!}{%
  {\small{
  \begin{tabular}{@{}c c|c c c @{}}
    \toprule
    \multicolumn{2}{c}{Method} & MVTec & VisA & Mean \\
    \midrule
    PatchCore & {\color{blue} 22'CVPR} & 50.74 & 28.15 & 39.44 \\
    DRAEM & {\color{blue}21'ICCV} & 27.93  & 17.35 & 22.64 \\
    DeSTSeg & {\color{blue}23'CVPR} & 43.61 & 23.48 & 33.55 \\
    RD4AD & {\color{blue}22'CVPR} & 39.73 & 8.34 & 24.02 \\
    UniAD & {\color{blue}22'NIPS} & 30.56 & 18.93 & 24.75 \\
    HVQ-Trans& {\color{blue}23'NIPS} & 32.63 & 26.71 & 29.67 \\
    \midrule
    UniAS (Ours)& {\color{blue}25'WACV} & \textbf{59.33} & \textbf{32.50} & \textbf{46.03}\\
    \bottomrule
  \end{tabular}
  }}}
  \vspace{-2mm}
  \caption{{Quantitative Results on MVTec-AD and VisA in DSC(\%)}. The best results are \textbf{bold}.}
  \label{tab:DSC}
  \vspace{-2mm}
\end{table}

As shown in \cref{tab:pAP} and \cref{tab:DSC}, our UniAS outperforms all the competitive models, setting new SOTA to $65.12/59.33$ and $40.06/32.50$ in the percentage of average pAP and DSC for two datasets, respectively. The performance has improved by a large margin, indicating the considerably improved segmentation ability of our UniAS. Specifically, for recent One-for-All SOTA models, namely UniAD and HVQ-Trans, feature downsampling in these methods damages the ability of anomaly localization and segmentation. Without this operation, UniAS outperforms the latest SOTA model HVQ-Trans~\cite{lu2023hierarchical}, by 16.97{$\color{red}\uparrow$}/26.70{$\color{red}\uparrow$} and 6.89{\color{red}$\uparrow$}/5.79{\color{red}$\uparrow$}. 
Moreover, the improvement compared to PatchCore~\cite{roth2022towards} indicates the effectiveness of utilizing the SAR learnable query as a memory matrix compared to a computationally expensive memory bank~\cite{roth2022towards}. 

\subsection{Ablation studies}
\begin{table}
  \centering
  \setlength{\tabcolsep}{ 3pt}
\resizebox{\columnwidth}{!}{%
  {\small{
  \begin{tabular}{@{}c|c|c c c|c c @{}}
    \toprule
    \multicolumn{5}{c|}{Components}& \multicolumn{2}{c}{Results} \\
    \midrule
    \multirow{2}{*}{Multi-level} & \multirow{2}{*}{SAR\ Q.} & \multicolumn{3}{c|}{Hybrid Structure} &\multicolumn{2}{c}{Metric} \\
    \cmidrule(lr){3-5}
    \cmidrule(lr){6-7}
    &  & Conv3 & MG-CNN & MGG-CNN & pAP & DSC\\
    \midrule
    - & -& -& -& - & 40.14{\color{red}(24.98)} & 25.24{\color{red}(34.09)}\\
    \checkmark & -& -& -& - & 56.47{\color{red}( 8.65)}  & 39.06 {\color{red}(20.27)} \\
    \checkmark & \checkmark & -& -& - & 59.84{\color{red}( 5.78)} & 48.35{\color{red}(10.98)}  \\
    \checkmark & \checkmark & \checkmark & -& - & 61.14{\color{red}( 1.73)} & 49.27{\color{red}(10.06)} \\
    \checkmark & \checkmark & -& \checkmark& - & 63.39{\color{red}( 1.13)} & 51.56{\color{red}( 7.77)}  \\
    \checkmark & \checkmark & -& -& \checkmark & \textbf{65.12} & \textbf{59.33}  \\
    \bottomrule
  \end{tabular}
  }}}
  \vspace{-3mm}
  \caption{\textbf{Structural Component Study} The performance gap from the default setting is shown in {\color{red}red}. The decoder of the baseline in the first line is a vanilla Transformer featuring self-attention layers exclusively. Cross-attention is employed to build a multi-level pipeline in the second line, with the deepest level still being formed by self-attention.}
  \label{tab:structure}
\end{table}
\begin{table}
  \centering
  \resizebox{0.75\columnwidth}{!}{%
  \small{
    \begin{tabular}{c|c|c c}
    \toprule
    \multirow{2}{*}{\#Levels} & \multirow{2}{*}{Levels} &  \multicolumn{2}{c}{Metric} \\
    \cmidrule(lr){3-4} 
    &  & pAP & DSC \\
    \midrule
    \multirow{4}{*}{1}&\{4\}& 44.44{\color{red}(20.68)} & 36.88{\color{red}(22.45)} \\
    &\{3\}& 53.49{\color{red}(11.63)} & 38.22{\color{red}(21.11)} \\
    &\{2\}& 56.52{\color{red}( 8.60)} & 37.47{\color{red}(21.86)} \\
    &\{1\}& 40.18{\color{red}(24.94)} & 21.11{\color{red}(38.22)} \\
    \midrule
    \multirow{2}{*}{2}&\{4,3\} & 54.20{\color{red}(10.92)} & 45.55{\color{red}(13.78)} \\
    &\{4,1\} & 58.63{\color{red}( 6.49)} & 45.63{\color{red}(13.70)} \\
    \midrule
    \multirow{1}{*}{3}&\{4,3,2\} & 61.57{\color{red}( 3.55)} & 55.03{\color{red}( 4.30)} \\
    \midrule
    4&\{4,3,2,1\} & \textbf{65.12} & \textbf{59.33} \\
    \bottomrule
    \end{tabular}
    }}
    \vspace{-3mm}
    \caption{\textbf{Feature Combination Study} The performance gap from the default setting is in {\color{red}(red)}. Feature levels are labeled 1,2,3,4 from lowest to highest level accordingly. \{$\cdot$\} means the levels included.}
  \label{tab:features}
\end{table}

\subsubsection{Structural Component Study}
In \cref{tab:structure}, we ablate each component of our structural designs on the MVTech-AD dataset: 

\noindent 1) \textbf{Multi-level}: Following UniAD~\cite{you2022unified}, the baseline model downsamples and reconstructs multi-level features at the lowest resolutions in the highest level. Compared to our UniAS with multi-level design, the performance experiences a notable drop of $24.98\downarrow/34.09\downarrow$. However, upon simply incorporating the multi-level reconstruction design, the performance improves significantly to $56.47/39.06$ in terms of pAP and DSC. This improvement underscores the complementary roles and necessity of multi-level features. 

\noindent 2) \textbf{Hybrid Structure}: We insert one layer of $Conv3$, Multi-granularity CNN (MG-CNN) without gate, and our MGG-CNN into Transformer layers. The results in the last three rows of \cref{tab:structure} show improvement against the third line, proving the importance of precisely modeling local details. Moreover, MG-CNN with gated branches of different receptive fields is more effective than a simple $Conv3$ layer, while the activation layer (``gate'') helps to aggregate multi-scale features, forming MGG-CNN.

\noindent 3) \textbf{Query}: We prove that simply incorporating a randomly initialized query for all datasets is less optimal in the One-for-All scheme. After deploying the SAR mechanism, significant performance improvements can be observed in the second line in ~\cref{tab:structure}.

\subsubsection{Feature Combination Study}
We confirm the essentiality and efficacy of amalgamating various feature levels, as presented in \cref{tab:features}. For the 4 levels of feature extracted by EfficientNet, we infer the trained model with various combinations of feature levels. It is observed that incorporating more levels leads to better performance. Although we discover that feature levels 2 and 3 contain more valuable information compared to levels 1 and 4, combining all feature levels leads to the best performance, which validates the effectiveness of our multi-level pipeline.

\section{Conclusion and Discussion}
Anomaly segmentation is critical for anomaly monitoring and removal but remains under-explored. Recent excellent AUROC results have obscured this issue in imbalanced AD settings. In this study, we carefully discuss the limitations of AUROC and introduce other metrics for anomaly segmentation, namely pAP and DSC. Moreover, we propose our model, UniAS, which advances anomaly segmentation within a class-unified framework. UniAS hierarchically reconstructs normal features through a multi-level hybrid pipeline, combining transformer and CNN layers, and leverages multi-granularity gated CNN to facilitate local multi-granular detailed filtering and reconstruction. Additionally, a sample-aware reweighting mechanism is integrated to enhance the robustness of One-for-All anomaly segmentation. Together, UniAS effectively discriminates defect-corrupted features at the pixel level, achieving new SOTA anomaly segmentation performance with significant improvements. 

Nevertheless, the performance of the proposed method in complex scenarios (\eg several classes in VisA) still offers potential for further enhancement. Challenges persist in accurately delineating the contours of anomalous regions, particularly in scenarios involving multiple objects or background noise. Segmenting anomalies accurately in intricate scenes remains an open challenge for future studies.
\section*{Acknowledgement}
This work is supported by Natural Science Foundation of China under Grant 62271465, Suzhou Basic Research Program under Grant SYG202338, Open Fund Project of Guangdong Academy of Medical Sciences, China (No. YKY-KF202206), and Jiangsu Province Science Foundation for Youths (NO. BK20240464).

{\small
\bibliographystyle{ieee_fullname}
\bibliography{egbib}
}

\end{document}


\title{Towards Accurate Unified Anomaly Segmentation\\ (Supplementary Material)} 

\maketitle
\vspace{-5mm}
\appendix

This Supplementary Material contains the following parts: 1) Additional information details about sample-aware reweighting mechanisms and hyper-parameter settings in \cref{sec:info}; 2) Additional experiments, ablation studies, and visualization results in \cref{sec:add-exp}.

\section{Implementation Details}
\label{sec:info}
\subsection{Sample-Aware Reweighting Mechanism}
{\small{
\begin{figure}[htbp]
    \centering
    \includegraphics[width=\columnwidth]{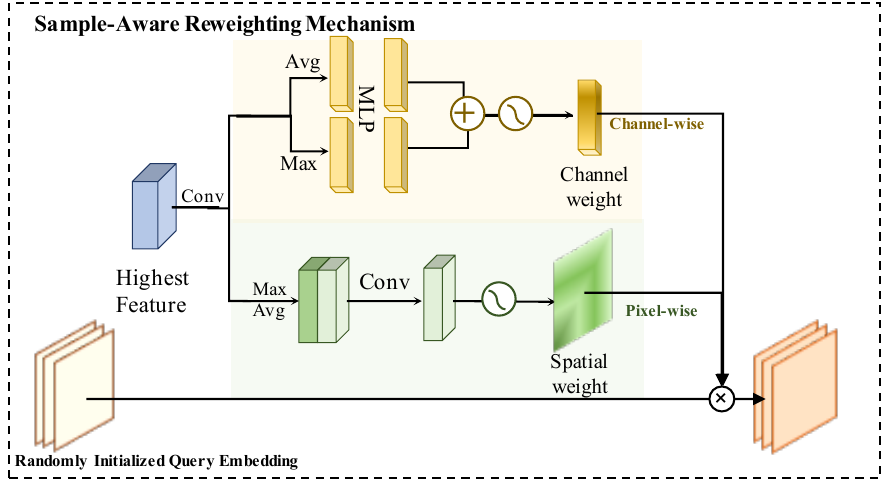}
    \caption{The illustration of the Sample-Aware Reweighting (SAR) Mechanism. Channel-wise attention and spatial-wise attention are computed based on the highest feature, which serves as the weights to reweight the randomly initialized query.}
    \label{fig:SARQ}
    \vspace{2mm}
\end{figure}}}
For One-for-All anomaly segmentation, a learnable query can be included to serve as the memory matrix, allowing for flexible memorization of class-agnostic semantics\cite{you2022unified, lu2023hierarchical}. However, features extracted from different samples, particularly those from distinct categories, often exhibit significant disparities in high-level characteristics. Employing a shared learnable query to memorize patterns across all categories can be suboptimal due to this variability. To address this issue, \cite{lu2023hierarchical} introduces a Switching Mechanism with various codebooks and experts, although this approach is memory-consuming.

Instead, we take advantage of this variability in various features, incorporating a simple Sample-Aware Reweighting (SAR) Mechanism to initialize more sample-specific accurate prototypes as queries, as illustrated in \cref{fig:SARQ}. Specifically, the query $\mathbf{q}_0^{ori} \in \mathbb{R}^{H_K \times W_K \times C'}$ is randomly initialized, whose shape is the same as the highest patch embedding $\mathbf{h}_K$.
We utilize channel weights $\mathbf{W}_c$ and spatial weights $\mathbf{W}_s$ in CBAM~\cite{woo2018cbam} to reweight query $\mathbf{q}_0^{ori}$ based on $\mathbf{h}_K$ with the richest semantic, as shown in \cref{eq:SARQ}. 
\begin{equation}
\label{eq:SARQ}
\begin{aligned}
\mathbf{W}_c &= \sigma(\text{MLP}(\text{AvgP}_s(\mathbf{h}_K)+\text{MaxP}_s(\mathbf{h}_K))),\\
\mathbf{W}_s &= \sigma(\text{Conv}(\text{cat}(\text{AvgP}_c(\mathbf{h}_K); \text{MaxP}_c(\mathbf{h}_K)))),
\end{aligned}
\end{equation}
where $\sigma(\cdot)$ is the activation function, $\text{AvgP}_s(\cdot),\text{MaxP}_s(\cdot)$ mean spatial-wise average-pooling and max-pooling, and $\text{AvgP}_c(\cdot),\text{MaxP}_c(\cdot)$ are channel-wise poolings.

The weights $\mathbf{W}_c, \mathbf{W}_s$ are channel-wisely and spatial-wisely multiplied to $\mathbf{q}_0^{ori}$, resulting in $\mathbf{q}_0 \in \mathbb{R}^{H_k \times W_k \times C'}$, which is the query of the first transformer layer.


\subsection{Hyper-parameter Settings}
The shapes of the four levels of features are $112\times 112\times 24$, $56\times 56\times 32$, $28\times 28\times 56$ and $14\times 14\times 160$, respectively. For the Gaussian filter, the kernel size is 3 and sigma equals 1. The patch size for each level is 8,4,2 and 1 accordingly. Inspired by~\cite{ding2022davit}, we combine two transformer components together in a transformer layer: a conventional spatial-wise transformer~\cite{vaswani2017attention} and a channel-wise transformer that performs attention operations on channels after transposing the input. This dual design enables the model to simultaneously consider spatial-wise context and channel-wise semantics, enhancing the model's performance and robustness in dealing with intricate scenarios across multiple datasets. For both spatial- and channel-wise attention, the number of heads is 4, and the dimension of the feed-forward network is 2048. All the channel sizes in the convolution layers in MGG-CNN module are set to 256. 

During evaluation, the threshold for segmentation is chosen based on the PR curve of each class to maximize the sum of precision and recall, and DSC is calculated sample-wisely. DSC for normal samples ($GT=0$) is calculated based on the following rules: 1) if each pixel in the predicted segmentation mask equals 0, then the DSC for this image is 1; 2) if any pixel in the segmentation mask is wrongly predicted, then the DSC for this image is 0.

\begin{table*}[h]
\centering
\vspace{3mm}
\resizebox{\textwidth}{!}{%
\begin{tabular}{l|c|c|c|c c|c c c}
\toprule
\multirow{3}{*}{Category} & {\color{blue}\multirow{3}{*}{AR(\%)}}  &Boundary&Img Recon.& \multicolumn{5}{c}{Feature Recon.}\\
\cmidrule(lr){3-3}\cmidrule(lr){4-4}\cmidrule(lr){5-9}\cmidrule(lr){5-9}
&  & PatchCore~\cite{roth2022towards} & DRAEM~\cite{zavrtanik2021draem}&DeSTSeg~\cite{zhang2023destseg} & RD4AD~\cite{deng2022anomaly} & UniAD~\cite{you2022unified} & HVQ-Trans~\cite{lu2023hierarchical} & \multirow{2}{*}{\textbf{UniAS(Ours)}} \\
&    & {\color{black}CVPR2022} & {\color{black}ICCV2021} & {\color{black}CVPR2023}  & {\color{black}CVPR2022} & {\color{black}NIPS2022} & {\color{black}NIPS2023} &  \\
\midrule
Candle & {\color{blue}0.81} & 13.21/\ 0.27  & 14.02/\ 0.27   & {\color{red}32.08}/\ 0.27 &  19.80/\ 0.00 & \underline{20.77}/\ 0.34     &  19.47/{\color{red}\ 5.38} & 19.09/\underline{\ 0.36} \\

Capsules & {\color{blue}1.63} & {\color{red}61.28}/{\color{red}51.78}   & 19.40/\ 0.77   & 22.47/\ 0.84 &  39.01/\ 0.00  & 49.22/45.71   &  45.98/37.79  & \underline{49.23}/\underline{46.43} \\

Cashew & {\color{blue}5.06} & \underline{54.81}/38.27  & \ 1.38/\ 1.84   & 51.89/{\color{red}47.04}  &  37.30/\ 0.01 & 42.91/\ 8.02 &  {\color{red}59.50}/36.01  & 52.43/\underline{38.64} \\

Chewing Gum & {\color{blue}2.54} & 45.53/51.74   & 43.68/45.78   &  \underline{61.59}/{\color{red}73.50}  &  62.09/32.22  & 57.97/\underline{61.22} &  47.70/55.71  & {\color{red}62.79}/54.32 \\

Fryum & {\color{blue}10.37} & 33.61/\ 5.92  & 34.03/\ 4.05   &  30.57/\ 3.99  &  {\color{red}51.46}/\ 0.07  & 46.87/21.02  &  \underline{50.16}/\underline{28.70} & 47.52/{\color{red}71.82} \\

Macaroni1 & {\color{blue}0.22} & \ 1.06/50.01  & \underline{14.82}/\ 0.07   &  \ 0.62/50.24  &  {\color{red}22.10}/\ 0.05  &\ 9.62/\ 0.03 &  \ 7.29/\underline{50.74}  & 14.73/{\color{red}50.78} \\

Macaroni2 & {\color{blue}0.17}& \ 0.03/\ 0.01  & \underline{10.42}/{\color{red}50.56}   & \ 6.01/\ 0.06  &  {\color{red}13.80}/\ 0.00  & \ 3.77/\ \underline{0.15}  &  \ 3.07/\ 0.14  & 0.17/\ 0.07 \\

Pcb1 & {\color{blue}2.79}& \underline{75.31}/\underline{58.44}  & 29.26/50.43   & 39.00/\ 0.88 &  74.84/46.63  & 70.01/56.81   &  59.98/55.57  & {\color{red}79.19}/{\color{red}58.58} \\

Pcb2 & {\color{blue}1.24}& \underline{18.09}/\ 0.05   & \ 7.76/\ 0.41   & 11.77/\ 0.53  &  {\color{red}18.93}/\ 0.00 & \ 9.67/{\color{red}\ 1.03}    &  \ 7.84/\ \underline{0.88}  & 11.95/ 0.69 \\

Pcb3 & {\color{blue}1.52}& {\color{red}30.35}/50.38  & 19.26/\underline{50.54}   & 26.16/{\color{red}50.67}  &  \underline{26.43}/\ 0.50  & 21.27/\ 1.59 &  18.59/\ 1.64  & 23.58/ 3.98 \\

Pcb4 & {\color{blue}3.96}& 36.17/\ 1.49   & 17.51/\ 1.32   & \underline{43.52}/\ 1.29  &  33.36/\ 0.40  & 29.72/\ 4.10  &  14.30/\ \underline{6.47}  &  {\color{red}44.47}/{\color{red}32.68} \\

Pipe Fryum & {\color{blue}5.62}& 58.23/44.57   & 27.78/\ 2.14   & \underline{74.52}/{\color{red}52.48}  &  59.70/20.74 & 50.04/26.89 &  64.17/46.28  & {\color{red}75.67}/\underline{51.23} \\

\midrule
Mean & {\color{blue}3.00}& 35.64/\underline{28.15} & 34.49/17.35  & 33.35/23.48  &  \underline{38.23}/\ 8.34 & 34.32/18.93  &  33.17/26.71  & {\color{red}40.06}/{\color{red}32.50} \\
\bottomrule
\end{tabular}
}
\caption{\textbf{Quantitative Results on VisA in pAP/DSC(\%)}. The best results are colored {\color{red}red}, and the second best results are \underline{underlined}. }
\vspace{3mm}
\label{tab:visa}
\end{table*}

\begin{table*}[h]
\centering
\vspace{3mm}
\resizebox{\textwidth}{!}{%
\begin{tabular}{l|c|c|c|c c|c c c}
\toprule 
\multirow{3}{*}{Category} & {\color{blue}\multirow{3}{*}{AR(\%)}}  &Mem. Bank & Img Recon.& \multicolumn{5}{c}{Feature Recon.}\\
\cmidrule(lr){3-3}\cmidrule(lr){4-4}\cmidrule(lr){5-9}
&    & PatchCore~\cite{roth2022towards} & DRAEM~\cite{zavrtanik2021draem} & DeSTSeg~\cite{zhang2023destseg}  & RD4AD~\cite{deng2022anomaly} & UniAD~\cite{you2022unified} & HVQ-Trans~\cite{lu2023hierarchical} & \multirow{2}{*}{\textbf{UniAS(Ours)}} \\
&    & {\color{black}CVPR2022} & {\color{black}ICCV2021} & {\color{black}CVPR2023}  & {\color{black}CVPR2022} & {\color{black}NIPS2022} & {\color{black}NIPS2023} &  \\
\midrule
Bottle & {\color{blue}22.82} & \underline{79.17}/\underline{78.42}  &  51.32/\ 1.04  &  72.93/65.14  & 68.07/47.39    & 69.34/67.72   &  72.02/65.86  & {\color{red}84.35}/{\color{red}79.50} \\

Cable & {\color{blue}14.04}& \underline{51.12}/\underline{51.36}  &  \ 9.53/\ 5.35   &47.50/46.35  & 25.68/\ 5.42   & 48.38/27.60  &  50.64/31.09  & {\color{red}78.31}/{\color{red}73.90} \\

Capsule & {\color{blue}3.32} & 44.26/\underline{43.41}  &  \ 7.15/\ 1.76 &\underline{45.88}/20.25 & 20.19/21.26    & 45.75/34.19 &  44.63/19.43  & {\color{red}49.64}/{\color{red}44.72} \\

Hazelnut & {\color{blue}10.06} & 60.12/\underline{63.23} &  \underline{71.82}/44.48  & 65.71/57.62 & 63.56/52.25  & 54.35/25.84 &  64.21/62.40  & {\color{red}79.17}/{\color{red}76.80} \\

Metal Nut & {\color{blue}43.46}& {\color{red}88.62}/\underline{74.35}  &  24.53/16.95   & 53.61/23.02  & 64.32/38.50   &49.83/41.30   &  67.30/48.18  & \underline{70.25}/{\color{red}76.61} \\

Pill & {\color{blue}11.93} & \underline{77.37}/\underline{51.25}  &  63.42/23.58  &{\color{red}78.03}/35.95  & 78.11/{\color{red}53.18}   &40.28/23.66  &  49.98/29.24  & 66.98/47.87 \\

Screw & {\color{blue}1.01}& 36.73/\ 0.00  &  41.49/\ 0.05  &19.39/\ 0.50  & \underline{43.89}/\underline{35.67}   & 26.08/\ 3.65  &  29.17/\ 3.57  & {\color{red}48.00}/{\color{red}46.54} \\

Toothbrush & {\color{blue}6.40} & 54.03/{\color{red}58.46}  & 52.60/34.09 &\underline{58.06}/43.04 & 55.51/53.96    & 40.09/42.81 &  39.80/45.56  & {\color{red}{59.41}}/\underline{55.42} \\

Transistor & {\color{blue}35.95} & 66.61/22.40  & 27.22/\ 7.07  & 39.26/\underline{59.02}  & 42.38/\ 8.05   & 67.57/28.87  &  \underline{72.34}/19.35  & {\color{red}84.13}/{\color{red}70.13} \\

Zipper & {\color{blue}7.87} & 53.50/53.79  & {\color{red}73.61}/\underline{57.93}  &\underline{61.42}/49.22 & 57.36/{\color{red}63.26}   &33.60/21.99 &  37.79/32.14  & 57.36/55.61 \\

\midrule
Carpet & {\color{blue}6.32}& 69.31/\underline{63.12}  & \underline{69.97}/52.34  & 66.82/52.52  & 57.49/48.81    &  52.81/49.74  &  54.19/47.49  &  {\color{red}70.20}/{\color{red}63.70}\\

Grid & {\color{blue}2.83} & 37.85/\underline{40.45}   & 38.49/17.50  &36.02/\ 1.36  & {\color{red}48.56}/{\color{red}43.69}   & 24.16/\ 3.80 &  24.09/\ 3.98 & \underline{43.00}/26.52 \\

Leather & {\color{blue}2.62}& 50.97/51.03  & \underline{58.99}/48.18  &{\color{red}79.37}/{\color{red}75.38}  & 40.98/45.44   &34.43/41.36  &  34.47/28.56  & 58.65/\underline{58.40} \\

Tile & {\color{blue}29.41}& 59.78/{\color{red}70.21}  & \underline{79.90}/45.77  &{\color{red}89.56}/\underline{60.65}  & 51.30/40.51   & 42.67/30.69  &  41.99/35.41  & 59.80/54.37 \\

Wood & {\color{blue}15.25}&  51.17/39.70  &  {\color{red}77.79}/\underline{62.81}  & \underline{74.57}/{\color{red}63.70}  & 53.69/38.53    &37.02/15.23    &  39.65/17.23  & 53.66/60.03 \\

\midrule
Mean & {\color{blue}13.17} &  55.73/\underline{50.74}  & 49.86/27.93  &\underline{59.20}/43.61  & 51.41/39.73 &44.42/30.56 &  48.15/32.63  & {\color{red}65.12}/{\color{red}59.33}\\
\bottomrule
\end{tabular}}
\caption{\textbf{Quantitative Results on MVTec-AD in pAP/DSC(\%)}. Methods are divided into Memory Bank-based, Image Reconstruction-based, and Feature Reconstruction-based categories. One-for-one methods are trained within the one-for-all scheme. Objects with logical and textural anomalies are separated. The best results are colored {\color{red}red}, and the second best results are \underline{underlined}.}
\label{tab:MVTec}

\end{table*}

\begin{table*}[ht]
  \centering
  \setlength{\tabcolsep}{3pt}
\resizebox{0.85\textwidth}{!}{%
  {\small{
  \begin{tabular}{@{}c|c|c c c|c c c|c c c@{}}
    \toprule
    \multicolumn{8}{c|}{Components} &\multicolumn{3}{c}{Results} \\
    \midrule
    \multirow{2}{*}{Multi-level} & \multirow{2}{*}{SAR\ Q.} & \multicolumn{3}{c|}{Hybrid Structure} &\multicolumn{3}{c|}{Filtering}&\multicolumn{3}{c}{Metric} \\
    \cmidrule(lr){3-5}
    \cmidrule(lr){6-8}
    \cmidrule(lr){9-11}
    &  & Conv3 & MG-CNN & MGG-CNN & Avg. & Gau. & Concat & pAP & DSC & AUROC\\
    \midrule
    - & -& -& -& - & -& \checkmark & \checkmark & 40.14{\color{red}(24.98)} & 25.24{\color{red}(34.09)} & 96.35{\color{red}( 1.87)}\\
    \checkmark & -& -& -& - & -& \checkmark & \checkmark & 56.47{\color{red}( 8.65)}  & 39.06 {\color{red}(20.27)} & 96.95{\color{red}( 1.27)} \\
    \checkmark & \checkmark & -& -& - & -& \checkmark & \checkmark & 59.84{\color{red}( 5.78)} & 48.35{\color{red}(10.98)} & 97.66{\color{red}( 0.56)} \\
    \checkmark & \checkmark & \checkmark & -& - & -& \checkmark & \checkmark & 61.14{\color{red}( 1.73)} & 49.27{\color{red}(10.06)} & 97.64{\color{red}( 0.58)}\\
    \checkmark & \checkmark & -& \checkmark& - & -& \checkmark & \checkmark & 63.39{\color{red}( 1.13)} & 51.56{\color{red}( 7.77)} & 98.09{\color{red}( 0.13)} \\
    \checkmark & \checkmark & -& - & \checkmark & -& - & - & 62.43{\color{red}( 2.69)} & 52.70{\color{red}( 6.63)} & 96.98{\color{red}( 1.24)} \\
    \checkmark & \checkmark & -& - & \checkmark & \checkmark & - & - & 62.79{\color{red}( 2.33)} & 51.77{\color{red}( 7.56)} & 98.03{\color{red}( 0.19)} \\
    \checkmark & \checkmark & -& - & \checkmark & - & \checkmark & - & 63.00{\color{red}( 2.12)} & 54.26{\color{red}( 5.07)} & 98.06{\color{red}( 0.16)} \\
    \checkmark & \checkmark & -& -& \checkmark & -& \checkmark & \checkmark & \textbf{65.12} & \textbf{59.33} & \textbf{98.22} \\
    \bottomrule
  \end{tabular}
  }}}
  \caption{\textbf{Structural Component Study} The performance gap from the default setting is shown in {\color{red}red}.}
\vspace{-3mm}
  \label{tab:ab-1}
\end{table*}

\begin{table}[t]
  \centering
  \resizebox{\columnwidth}{!}{%
  \small{
    \begin{tabular}{c|c|c c c}
    \toprule
    \multirow{2}{*}{\#Levels} & \multirow{2}{*}{Levels} &  \multicolumn{3}{c}{Metric} \\
    \cmidrule(lr){3-5} 
    &  & pAP & DSC & AUROC\\
    \midrule
    \multirow{4}{*}{1}&\{4\}& 44.44{\color{red}(20.68)} & 36.88{\color{red}(22.45)} & 94.83{\color{red}( 3.39)} \\
    &\{3\}& 53.49{\color{red}(11.63)} & 38.22{\color{red}(21.11)} & 96.37{\color{red}( 1.85)} \\
    &\{2\}& 56.52{\color{red}( 8.60)} & 37.47{\color{red}(21.86)} & 96.16{\color{red}( 2.06)} \\
    &\{1\}& 40.18{\color{red}(24.94)} & 21.11{\color{red}(38.22)} & 89.30{\color{red}( 8.92)}\\
    \midrule
    \multirow{2}{*}{2}&\{4,3\} & 54.20{\color{red}(10.92)} & 45.55{\color{red}(13.78)} & 96.85{\color{red}( 1.37)} \\
    &\{4,1\} & 58.63{\color{red}( 6.49)} & 45.63{\color{red}(13.70)} & 97.45{\color{red}( 0.77)}\\
    \midrule
    \multirow{1}{*}{3}&\{4,3,2\} & 61.57{\color{red}( 3.55)} & 55.03{\color{red}( 4.30)} & 97.85{\color{red}( 0.37)} \\
    \midrule
    4&\{4,3,2,1\} & \textbf{65.12} & \textbf{59.33} & \textbf{98.22} \\
    \bottomrule
    \end{tabular}
    }}
    \caption{\textbf{Feature Combination Study} The performance gap from the default setting is in {\color{red}(red)}. Feature levels are labeled 1,2,3,4 from lowest to highest level accordingly. \{$\cdot$\} means the levels included.}
  \label{tab:ab-2}
  \vspace{-3mm}
\end{table}

\section{More Experiments}
\label{sec:add-exp}
\subsection{Results per Class}
We provide the the results of each class in our experiment on MVTec-AD (see \cref{tab:MVTec}) and VisA (see \cref{tab:visa}), comparing with exisiting SOTA models. We also present {\color{blue}AR} to reference the degree of imbalance in the dataset.

\subsection{Ablations}
We provide ablation results in pAP, DSC, and AUROC in this section, see \cref{tab:ab-1} for Structural Component Study and \cref{tab:ab-2} for Feature Combination Study.

We also provide additional ablation results of feature filtering. Although the effectiveness of aggregating neighbor information has been validated by previous works, we analyze the importance of our Gaussian filter in our settings, shown in \cref{tab:ab-1}. Average filters (Avg. column in \cref{tab:ab-1}) are used in previous works~\cite{roth2022towards, liu2023simplenet}, which are proved to be too smooth to reserve necessary structural information compared to the Gaussian filter we use (Gau. column in \cref{tab:ab-1}) , according to the last four lines. Moreover, concatenating (Concat column in \cref{tab:ab-1}) the residual together with filtered features improves segmentation performance significantly, showing that incorporating details is beneficial to fine-grained localization (see last two lines).

\subsection{Visualization}
{\small{
\begin{figure}[h]
    \centering
    \vspace{-2mm}
    \includegraphics[width=\columnwidth]{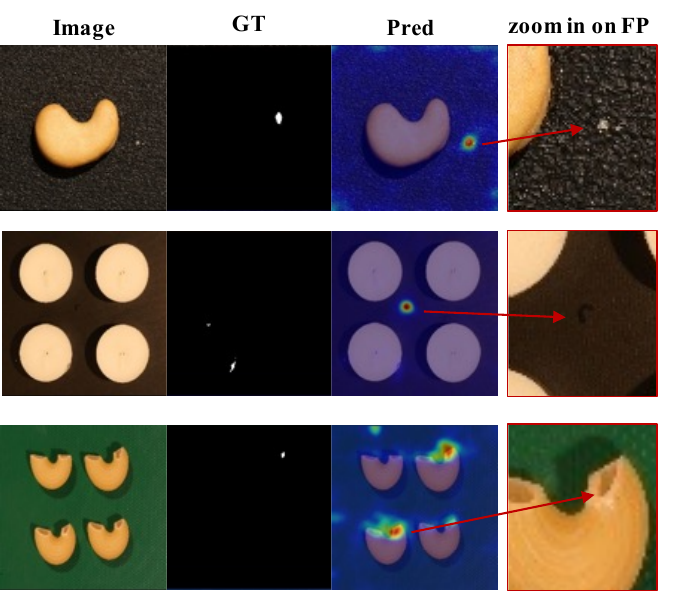}
    \caption{Examples of failed cases of our UniAS. Our model can make false predictions under noisy and complicated scenarios.}
    \label{fig:failed}
    \vspace{-3mm}
\end{figure}}}

{\small{
\begin{figure*}[h]
    \centering
    \includegraphics[width=0.86\textwidth]{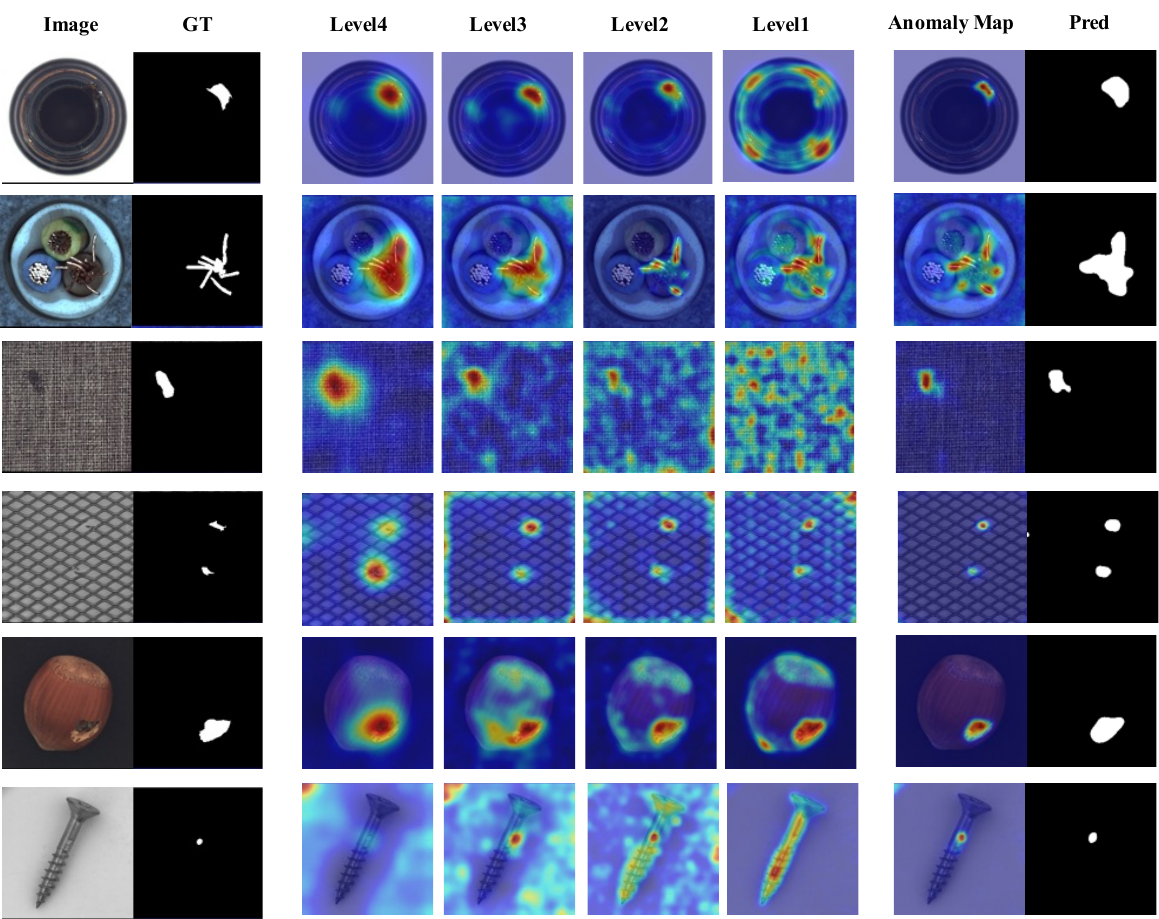}
    \caption{Additional visualization of the anomaly maps in each level and final predictions. UniAS leverages the benefits of multi-level reconstruction and achieves meaningful segmentation of anomaly.}
    \label{fig:vis}
    \vspace{-3mm}
\end{figure*}
}}
\subsubsection{Multi-level Reconstruction}
We visualize more examples of UniAS's prediction on every level in \cref{fig:vis}, labeled 4 to 1 from the highest to the lowest. These visualization result further consolidates that anomaly maps in different levels play complementary roles, detecting anomalous pixels from course to fine. Higher-level features concentrate on semantics, facilitating accurate overall anomaly localization, while lower-level features have rich textural information aiding in delineating the shape of the anomalous region.

\subsubsection{Failure Cases}
We show some failed cases of UniAS in \cref{fig:failed}. As one can see, when the anomalous region is vague while there is obvious background noise (seen in the first two lines), our model can possibly recognize the background noise as anomaly with greater salience than the real anomaly in the foreground. Additionally, there are some noisy labels in the datatset as well (seen in the third line). This leads to false positive predictions, implying our UniAS still needs further improvement, especially under intricate and confusing situations. However, anomaly segmentation with noise is a slightly different task, which is not the topic of our work. 

\newpage
{\small
\bibliographystyle{ieee_fullname}
\bibliography{egbib}
}